\documentclass[9pt]{IEEEtran}
\usepackage[letterpaper,%
            left=0.75in,right=0.75in,top=1in,bottom=1in,%
            footskip=1in]{geometry}
\usepackage{amsmath,amssymb,amsfonts}
\usepackage{mathtools}
\usepackage{algorithmic}
\usepackage{graphicx}
 \usepackage{caption}
\usepackage{booktabs}
\usepackage{array}
\usepackage{textcomp}
\usepackage{xcolor}
\usepackage{tabularx}
\usepackage{hyperref}

\usepackage[numbers,sort&compress]{natbib}

\newcommand{\argmax}{\arg\!\max}
\usepackage{textcomp}
\usepackage{xcolor}

\title{Deep Reinforcement Learning for Adaptive Exploration of Unknown Environments}

\author{\IEEEauthorblockN{Ashley Peake, Joe McCalmon, Yixin Zhang, Daniel Myers, Sarra Alqahtani, Paul Pauca}\\
\IEEEauthorblockA{\textit{Computer Science Department} \\
\textit{Wake Forest University}\\
Winston-Salem, North Carolina, USA \\
\{peakaa19, mccajl18, zhany217, myerdc18, sarra-alqahtani, paucavp\}@wfu.edu}
}

\begin{document}


\maketitle

\begin{abstract}
Performing autonomous exploration is essential for unmanned aerial vehicles (UAVs) operating in unknown environments. Often, these missions start with building a map for the environment via pure exploration and subsequently using (i.e. exploiting) the generated map for downstream navigation tasks. Accomplishing these navigation tasks in two separate steps is not always possible or even disadvantageous for UAVs deployed in outdoor and dynamically changing environments. Current exploration approaches either use a priori human-generated maps or use heuristics such as frontier-based exploration. Other approaches use learning but focus only on learning policies for specific tasks by either using sample inefficient random exploration or by making impractical assumptions about full map availability. In this paper, we develop an adaptive exploration approach to trade off between exploration and exploitation in one single step for UAVs searching for areas of interest (AoIs) in unknown environments using Deep Reinforcement Learning (DRL). The proposed approach uses a map segmentation technique to decompose the environment map into smaller, tractable maps. Then, a simple information gain function is repeatedly computed to determine the best target region to search during each iteration of the process. DDQN and A2C algorithms are extended with a stack of LSTM layers and trained to generate optimal policies for the exploration and exploitation, respectively. We tested our approach in 3 different tasks against 4 baselines. The results demonstrate that our proposed approach is capable of navigating through randomly generated environments and covering more AoI in less time steps compared to the baselines. 


\end{abstract}




         



\section{Introduction}

Exploration of an unknown environment is an important task in many applications of mobile robotics. In large, outdoor environments, Unmanned Aerial Vehicles (UAVs) are often employed for exploration due to their ease of use and maneuverability. Typically, a map of the target environment is built (via Simultaneous Localization and Mapping, i.e. SLAM) and is subsequently exploited in downstream navigation tasks~\cite{chen2019learning}. Accomplishing these navigation tasks in two separate steps is not always possible in dynamically changing environments. To overcome this, \textit{adaptive exploration} can be used to combine and trade off between exploration and exploitation in one single step, allowing the UAV to efficiently collect relevant information from the environment. Developing adaptive exploration algorithms is particularly important in time-critical tasks in outdoor environments such as Search and Rescue (SaR)~\cite{Sampedro}, illegal activity detection (e.g. human-trafficking, gold mining \cite{liu2019robustnessdriven}), or environmental applications (localizing wildlife, pollution, or mobile machinery). 

Reinforcement Learning (RL) has been utilized to learn exploration policies in training environments \cite{chen2019learning,Sampedro,Bayerlein,Imanberdiyev,Pham}.
RL algorithms do not require the agents to have an explicit model of the environment to properly navigate within it. Instead, the RL agent learns how to act in an environment by repetitively interacting with it and receiving positive or negative feedback from a predefined reward function~\cite{Sutton}. Moreover, RL algorithms generalize well to novel environments when combined with deep learning networks and can greatly adapt to new scenes in testing environments. In this paper, we propose a novel approach for adaptive exploration that simultaneously prioritizes coverage sampling and area of interest (AoI) sampling, without depending on user-defined weighted objectives as in ~\cite{10.1145/1273496.1273553,10.5555/1402383.1402392,10.1007/978-3-662-44845-8_43,6907763}. We use deep reinforcement learning (DRL) to learn an adaptive exploration policy $\pi_{AE}$ that allows the agent to autonomously explore an unknown environment while prioritizing navigation of AoIs. There are, however, significant challenges to overcome. First, outdoor environments can be theoretically extended infinitely in each direction. A large region to explore introduces a large set of possible observations that the RL agent (i.e. UAV) can encounter. While deep learning networks can greatly adapt to the unseen observations, performance will still be hindered, and the training process will be expensive. Additionally, in such environments, to take advantage of previously discovered information, each individual observation may include thousands of features and can quickly make the problem intractable. Second, RL algorithms are known to have poor convergence in partially observable environments. Since the agent can only observe a limited area accessible by its on-board camera, the search environment becomes partially observable for the agent at any time step. Lastly, the RL agent in this problem has no predefined, stable condition to end a training episode, i.e.~reaching a predefined target location, which can make it very hard for the RL agent to train on any reward function. To the best of our knowledge, this paper is the first to address the exploration problem with those constraints.

To tackle these challenges, we design an adaptive exploration policy architecture. First, we employ a map segmentation technique to decompose the environment into disjoint regions, where each region contains the probability that the region contains an AoI, provided through exploration. This allows the UAV to only navigate within a small region at a time, while still preserving information from across the entire mission for use in decision making (i.e. egocentric exploration). We then train a Deep Reinforcement Learning (DRL) model to direct UAV flight to optimize efficient information gain. We incorporate two neural networks, each selecting an action according to one of two competing navigation tasks. On one hand, the UAV must quickly explore the entire environment in search of AoIs. On the other, it must fully cover or "exploit" AoIs that have already been located (i.e. image or deploy sensors for detailed information collection). We call these the exploration and exploitation tasks, respectively. To control the trade-off between these tasks, we design an efficient information gain function that utilizes current knowledge the UAV has collected about the entire map of the environment (i.e. allocentric exploration).

To train the exploration task, we use the double deep Q-Learning (DDQN) algorithm~\cite{Hausknecht} to generate an optimal flight trajectory to a target region, while still prioritizing flying over AoIs on the path to that target. For AoI exploitation, we train another DRL model using the Actor-Critic (A2C) algorithm~\cite{Sutton} that allows the UAV to freely explore its current region in search for AoIs. To deal with the poor convergence of RL algorithms in partially observed environments, we embed a stack of recurrent layers, in particular long-short term memory (LSTM), to the DDQN and A2C models. The LSTM layers combat partial observability by letting the UAV maintain an internal representation of the areas it has visited, despite not having direct access to those state representations.

We have tested our approach in simulation on two different settings: 1) $35$ randomly simulated maps of AoIs of different shapes and distributions and 2) $19$ maps from Planet satellite imagery for detecting illegal gold mining activity in Amazonian forests in Peru~\cite{liu2019robustnessdriven}. We built a random scenario generator to create the training and testing maps for the first task. The maps are structured as $180 \times 180$ cell grids. The scenario generator randomly places the AoI using a normal distribution with values of the standard deviation and mean based on map width and height. Additionally, we tested our approach in hardware using a Pixhawk quadcopter connected to a Raspberry Pi 3 Model B to search for specific targets in a grassy field. We compared the results of our approach against several baselines and conducted ablation experiments to identify which parts of our proposed solution contribute the most to the performance.  Overall, the results reveal that our model is able to more efficiently cover AoIs in an unknown environment compared to the baselines. For reproducibility, our code can be accessed through \url{https://github.com/RL-WFU/Drone_field.git}

The remainder of this paper is organized as follows. We discuss previous and related work on autonomous controllers for UAVs in Section~\ref{sec:relatedWork} and an overview of RL and other algorithms used in Section~\ref{sec:rl}. Section~\ref{sec:adaptiveExploration} formulates the problem and introduces the proposed approach. Section~\ref{sec:experiments} presents our experiments and discusses results. Final remarks and conclusions are offered in Section~\ref{sec:conclusion}.

\section{Related Work} \label{sec:relatedWork}
As we study the problem of trading off between exploration and exploitation in navigation tasks, we draw upon recent efforts that use heuristic-based and learning-based approaches for this problem. We survey related efforts in these two directions.

\subsection{Heuristic-Based Exploration}
Early autonomous exploration methods explored simple environments, for example, by following walls or similar obstacles. Frontier exploration~\cite{613851} was one of the the first exploration methods capable of exploring a generic 2D environment. Frontier regions are defined as the borders between free and unexplored areas. Exploration is done by sequentially navigating close frontiers. Advanced variants~\cite{6697120,7989177} of this algorithm improve the coverage of unknown space along the path to the frontier. 

Next-best-view (NBV) exploration methods are a common alternative to frontier-based exploration. Recently, a Receding Horizon NBV planner~\cite{7487281} was proposed for online autonomous exploration of unknown 3D spaces. This planner employed rapidly exploring random trees (RRT) with a cost function that considered the information gain at each node of the tree. A path to the best node was extracted and the algorithm was repeated each time the vehicle moved along the first edge of the best path. An extension of this work~\cite{8633925} resolves the problem of sticking to local minima by extending it with a frontier-based planner for global exploration. In our work, we choose to use RL over RRT and its extensions for two reasons. First, RRT typically employs a uniform proposal distribution for sampling which does not make use of the structures of the environment and thus may require many samples to obtain an initial feasible solution path in a new environment. Second, the RRT does not have any systematic way to take advantage of information from previous experiences. It requires computing samples from scratch to build trees whenever a new start or goal configuration is specified, even when a similar solution has been computed on a previous query in that environment, which is what RL excels in. 

Temporal logic has also been used in the context of robotic motion and path planning in unknown environments. For instance, deterministic $\mu$-calculus was used to define specifications for sampling-based algorithms~\cite{8754788}. 
Robustness of Metric Temporal Logic (MTL) has been embedded in A* to increase the safety of UAVs navigating adversarial environments~\cite{Alqahtani}. In~\cite{liu2019robustnessdriven}, the NBV is sampled according to the current field of view of the UAV. Here, the views are randomly sampled as potential targets via Markov Chain Monte Carlo methods and evaluated by their robustness values of the probabilistic metric temporal logic (P-MTL) constraints. In Section~\ref{sec:experiments}, we compare our approach against the P-MTL method~\cite{liu2019robustnessdriven} for the illegal mining search task.

\subsection{Learning-Based Exploration}
Recently, machine learning has been used to develop autonomous navigation and exploration algorithms for robotics~\cite{Bayerlein, Imanberdiyev,chen2019learning,Pham,Sampedro}. A number of design choices have been investigated including different policy architectures for representing the exploration space. For example, \cite{zhu2016targetdriven} uses feed-forward networks, \cite{mirowski2017learning} uses vanilla neural network memory, \cite{Gandhi} uses spatial memory and planning modules, and \cite{savinov2018semiparametric} uses semi-parametric topological memory. In \cite{Imanberdiyev}, a model-based RL algorithm is used for autonomous navigation that learns faster than traditional table-based Q-learning methods due to its parallel architecture. In \cite{Pham}, a function approximation-based RL approach is exploited for a large number of states. The use of function approximation reduces the convergence time of the algorithm needed for search and rescue operations. A deep deterministic policy gradient is developed in \cite{Gandhi} to provide results in continuous time and action. The use of neural networks and deep learning was briefly performed in \cite{Lygouras} to guarantee safe take-off, navigation, and landing of the UAV on a fixed target.

Instead of focusing only on navigation for a specific task or on pure exploration of an environment, however, here we combine both to efficiently train our adaptive exploration policy. We achieve this by encouraging the RL agent to actively exploit the sensed information in its egocentric map via the exploitation task, while escaping local optima by prioritizing exploring unseen areas in the allocentric map via the exploration task. In doing so, unlike previous work, we do not assume access to human demonstrations in the given novel training and test environments~\cite{savinov2018semiparametric,chen2019learning}, nor do we assume availability of prior knowledge about the environment~\cite{zhu2016targetdriven,mirowski2017learning}. Our RL agent starts with no knowledge about the environment and derives it using the on-board camera during runtime. This makes our proposed approach amenable to real world deployment as shown in Section~\ref{sec:experiments}. 

\section{Reinforcement Learning} \label{sec:rl}
Reinforcement learning is suitable for solving problems that can be formulated as a Markov decision process (MDP). An MDP is a method for framing an environment such that it can be written as a tuple containing the state space $S$, the action space $A$, the reward space $R$, and the probability of transitioning from a state $s_t$ at time $t$ to another state $s_{t+1}$ at time $t+1$. An MDP is defined formally as $p(r_{t+1}, s_{t+1}| a_{t}, s_{t})$, for all time steps $t$, and additionally must satisfy the Markov property. This property necessitates that this transition only depends on the state-action pair at the current time step $t$, and not on any state-action pairs from prior time steps. 

Actions are chosen by the RL agent according to its policy, $\pi(a_{t} | s_{t}) = p(a_{t} | s_{t})$. The optimal policy is one which maximizes the state-action value function: 
\begin{equation} \label {eq:Qsa}
    Q_{\pi}(s_{t}, a_{t}) = \mathbb{E} \left[\sum^{\infty}_{\ell} \gamma^{\ell} R_{t+\ell+1}|S_{t}=s_{t}, A_{t}=a_{t} \right ]
\end{equation}
where $\gamma$ is a discount factor between $0$ and $1$. $R_{t}$ is the expected reward from a state-action pair,$(s_{t}, a_{t})$. For $\ell$, we have the expected reward at time instant $t+1$:
\begin{equation} \label{eq:expReward}
\begin{aligned}
    r_{t+1}(s_{t},a_{t})&=\mathbb{E} [R_{t+1}|S_{t}=s_{t},A_{t}=a_{t}]
   \\& = \sum_{r_{t+1}\in \mathbb{R}} r_{t+1} \sum_{s_{t+1}\in S} p(r_{t+1},S_{t+1}|S_{t},a_{t})
\end{aligned}    
\end{equation}

The goal of the RL agent is to maximize expected future reward. If the transitions from Eq.~\ref{eq:Qsa} are known, then an MDP of this nature can be solved through policy evaluation and policy iteration. However in most practical cases, the exact transition probabilities are unknown, so the agent must collect experience by interacting with the environment and internally computing the state-action value function itself. This process can be done with tabular methods in small environments, such as in Q-learning, or with a neural network such as in deep Q-learning (DQN). In the case that the Markov property is not satisfied, and the RL agent does not have full access to the environment at any given time step, the environment can be framed as a partially observable Markov decision process (POMDP). By using LSTM layers in the agent’s policy and value neural networks, the agent can maintain an internal representation of the entire environment, despite only observing part of it, in order to more closely resemble an MDP. In this paper, we use the double DQN (DDQN) and advantage-actor critic (A2C) algorithms to train the exploration and exploitation tasks respectively while enhancing their networks with a stack of LSTM layers to deal with limited observability. We use DDQN for training in exploration due to its ability to learn to reach target locations and its strength at reducing the overestimation of the values of successor states~\cite{hasselt2015deep}. A2C is used in exploitation training because of its ability to learn a successful policy in continuous, rather than episodic, environments~\cite{Sutton}.

\section{Adaptive Exploration Policy for Unknown Environments} \label{sec:adaptiveExploration}
In this paper, the agent's objective is to learn an adaptive exploration policy $\pi_{AE}$ enabling it to efficiently explore the given environment while prioritizing navigating detected AoIs. We formulate the estimation of $\pi_{AE}$ as a learning problem. We design a reward function that is estimated using the features extracted from the environment. $\pi_{AE}$ is learned on a set of training environments and tested on a different set.

Adaptive exploration in outdoor, large environments requires long-term coherent behavior over multiple time steps, such as avoiding obstacles, exploring new directions, finding AoIs and following them. Such long-term behavior is hard to learn using purely RL, given sparsity in reward which leads to excessively large sample complexity for learning. To overcome this large sample complexity, we design a simple map segmentation technique to make learning more tractable for the exploration and exploitation policies. 


\subsection{Map Segmentation for Tractable Learning}
Given an unknown, outdoor environment $E$, we first structurally decompose the continuous space into a regular grid of smaller regions. We divide the map of $M$ of size $W \times H$ into $N$ regions of size $\frac{W \times H}{\alpha}$ where $\alpha$ is a hyperparameter to control the learning performance of the agent. Thus, each region $R$ gives a large enough map for the agent to navigate, without being so large that learning is intractable. We then dynamically create two maps with different levels of egocentrism around the agent. One is a detailed map $o_t$ representing the agent's current observation of dimension $w \times h$, determined by the camera vision range. The second is a course map $m_t$ of size $w^{2} \times h^{2}$ that contains information collected about the area around the agent. This design choice is inspired by \cite{chen2019learning,parisotto2017neural,8578982}, though our work improves upon the levels of egocentrism, increasing the success of $\pi_{AE}$ as we will show in Section~\ref{sec:experiments}.

 The agent updates $m_t$ based on its current observation while navigating through the area. When the agent moves outside the bounds of $m_t$, $m_{t+1}$ will be centered around the agent's new position. This egocentric map $m_t$ is then used to update the allocentric map $M$, which stores the probability distribution of AoI in each region. We similarly record encoded information for each cell based on whether it has been visited by the agent. This value is scaled if the agent has visited the cell more than once. We use this to construct a vector $V$ of the visited information for each adjacent cell to the UAVs current position in order to enhance the learning of $\pi_{AE}$ (next section).

\begin{figure}[h]
\centering
\includegraphics [width=0.45\textwidth]{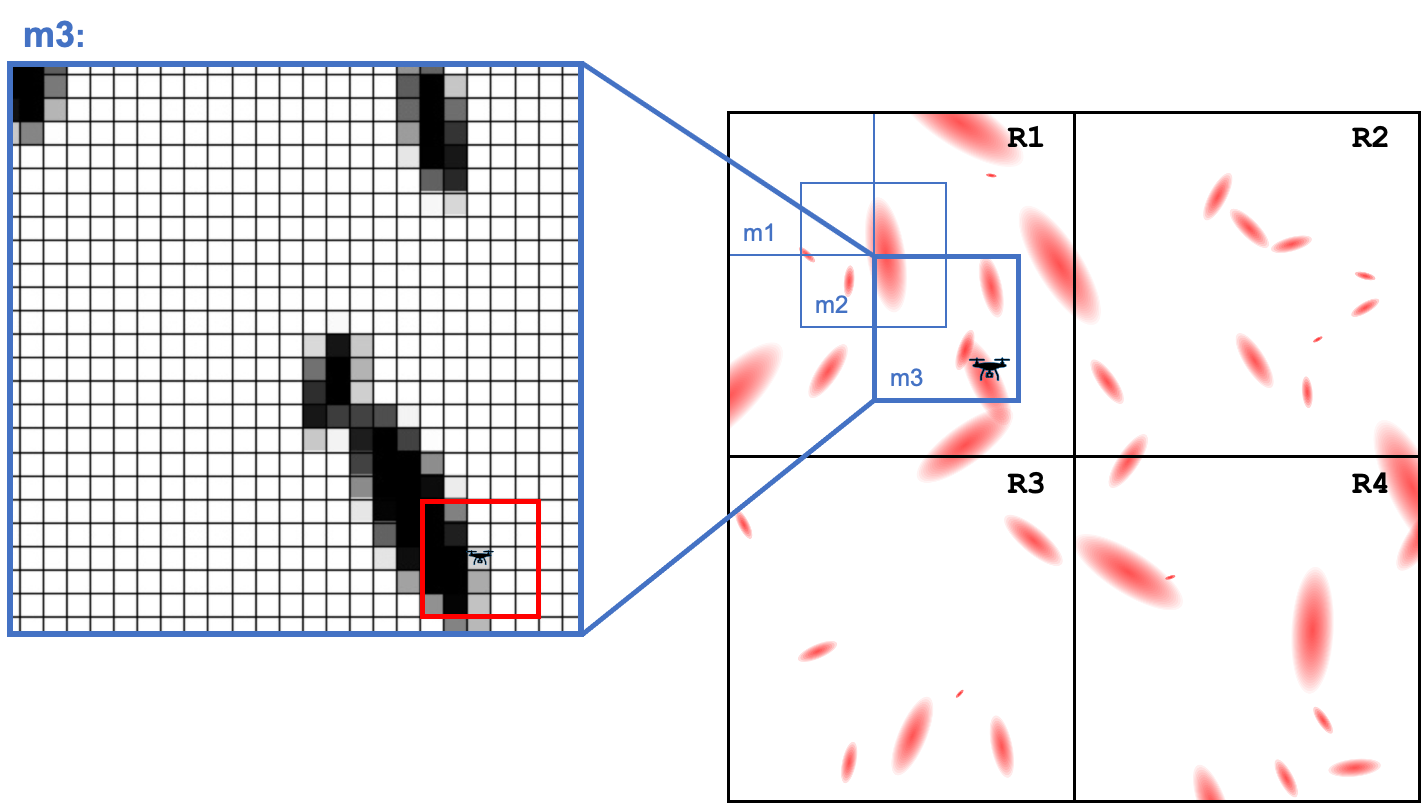}
\caption{Abstraction of the environment map into regions. Red areas on the map are simulated AoIs. The blue squares represent the egocentric maps $m_t$. The enlarged square shows the AoI distribution information stored in the map $m_t$. The red square around the agent represents its current observation $o_t$.}
\label{fig:quad}
\end{figure}

\subsection{Adaptive Exploration Policy Architecture}
\begin{figure*}
  \centering
  \includegraphics[scale=.4]{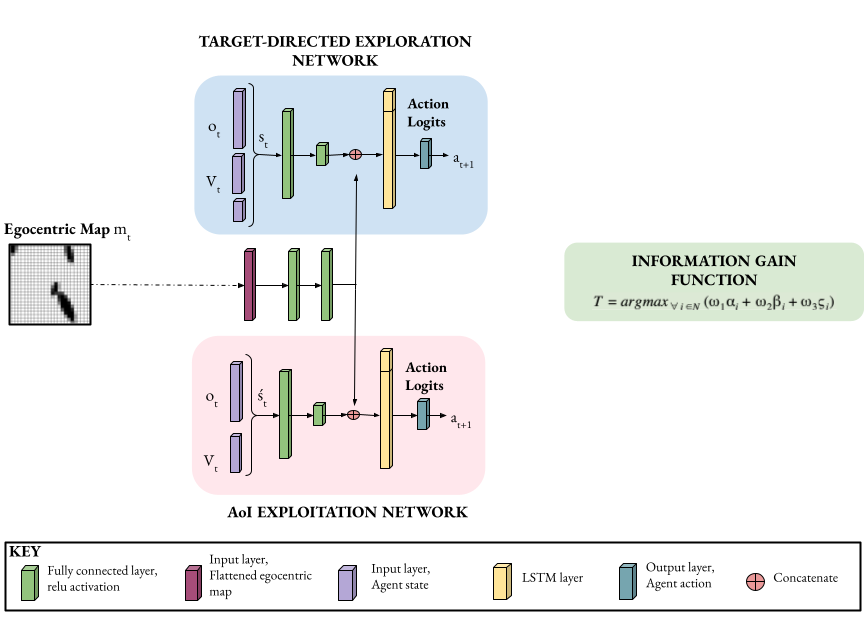}
  \caption{ \label {fig:network} The network architecture for training Recurrent-DDQN and Recurrent-A2C in Target-Directed Exploration and Region Exploitation. The information gain function determines which network to prioritize at time step $t$.}
 
\end{figure*}

We design the learning architecture of the adaptive exploration policy $\pi_{AE}$ in two sub-policies: one policy for Target-Directed Exploration $\pi_{n}$ and the other for AoI Exploitation $\pi_{r}$. Decomposing the policy into 2 sub-policies improves the efficiency of learning process of $\pi_{AE}$ and enables the generated policy to successfully complete the exploration task. The policy architecture is shown in Fig.\ref{fig:network} and described in more detail in the following sections.
\subsubsection{ Target-Directed Exploration Policy $\pi_{n}$ }
Under policy $\pi_n$ the agent prioritizes exploration across the entire search space by navigating to the next best region $R$. The agent determines the best region to navigate by using the allocentric map of the environment $M$ to compute the information gain from each region. A simple objective function (Eq.\ref{eq:targets}) is then used to identify the region $i\in N$ with the greatest potential for having unexplored AoI based on: the percentage of the region which the agent has previously visited $(\alpha_{i})$, the concentration of AoI previously seen in the region $(\beta_{i})$, and the Euclidean distance from the region's centroid to the current location of the agent $(\zeta_{i})$. 

\begin{equation} \label{eq:targets}
    T= \argmax_{\forall i \in N}\; (\omega_{1}\alpha_{i}\;+ \omega_{2} \beta_{i} \;+ \omega_{3} \zeta_{i}) 
\end{equation}

where $\omega_{1}, \omega_{2}, \omega_{3}$ represent the weights for coverage, AoI probability, and the distance from the agent to the region $i$, respectively. It is worth mentioning that the coverage and distance are weighted with negative $\omega$ to represent the cost, while the weight of the potential AoI is positive to represent the information gain.

To navigate to the assigned region $R$, the agent learns to maximize the AoI coverage over its flight trajectory, while efficiently reaching $R$. The agent learns its $\pi_{n}$ by fusing information from its current observation $o_t$, its current egocentric map of $m_t$, and the recently updated visited vector $V_t$: 


\begin{enumerate}
\item {\textbf{Observation $o_t$:} The images taken by the agent's camera are processed through any classification or detection algorithm such as CNN to extract and detect the features of AoI. In our simulated experiment, we use a simple color thresholding technique to detect AoI. For our gold mining experiment, we use a CNN to recognize the mining machines and traces as AoI based on a dataset of remote sensing images taken from Planet Satellite and UAVs as in \cite{liu2019robustnessdriven}.}
\item {\textbf{Egocentric Map $m_t$:} We use previous observations up to time step $t$ to derive the egocentric map $m_t$. We use $m_t$ instead of using the allocentric map $M$ to simplify the learning process by using only information local to the current position of the agent $\hat {x_t}$. This map allows the navigation algorithm to not only utilize previously detected AoI but also to locate it with respect to $\hat {x_t}$.}
\item {\textbf{Visited Vector $V_t$:} We use information about visited and unvisited cells in $m_t$. This information is scaled based on the number of times a cell has been visited, and is stored across the entire map, effectively creating a heat map. Local visited information helps the agent explore areas it has not been to before, which eventually increases the potential of covering more AoI.}
\end{enumerate}

The Target-Directed Exploration policy $\pi_n$ is trained by the Recurrent-DDQN algorithm. The agent navigates within its current $m_t$ using $\pi_{n}$ until it reaches the cell closest to the target region, denoted $T_L$. Once the agent reaches $T_L$, a new $m_{t+1}$ is created with the agent’s current position at the center and a new $T_L$ is selected accordingly.
The state $s_t$ is the first input into the Recurrent-DDQN network at each time step. $s$ is a vector of flattened information from $o_{t},V_{t}$ and the distance between the agent and the assigned destination. The first $h \times w$ entries are the probabilities of detecting an AoI in the $o_t$. The next 4 entries represent information about how frequently the agent has visited the adjacent cells (left, right, top, bottom), retrieved from $V_t$ in the allocentric map $M$. $V(x,y) \in (0,1)$ if the location $(x,y)$ has been visited by the agent, with a value closer to 1 corresponding to a higher number of visits to that cell. Otherwise, if the cell has not been previously visited, it has value of 0. The last two entries in the state $s$'s vector are the Euclidean distance in the $x$ and $y$ directions to $T_L$. (See Fig.\ref{fig:network}).

The second input to the Target-Directed Navigation network is the agent's current egocentric map $m_t$. The state $s_t$ and egocentric map $m_t$ are then processed in the network to select the optimal action for the agent. As the agent observes new information in its $o_t$, the probabilities of detecting AoI are added to the allocentric map, $M$, which contains all the information that has been discovered throughout the entire training episode.


\subsubsection{AoI Exploitation $\pi_{r}$}

Target-Directed Exploration is prioritized until the agent reaches its assigned target region $R$, or when the step limit of $\kappa_1$ is reached. Then, the agent transitions to the AoI Exploitation policy $\pi_r$. Given an Area of Interest (AoI) of unknown size and distribution within a region of the environment, $\pi_r$ aims to maximize AoI coverage while minimizing the number of steps taken without a specific target to reach. This policy uses the same $m_t$ architecture as $\pi_n$. The dynamics of $m_t$, however, differ slightly between the two policies. Since the goal of AoI exploitation is to maximize the probability of detecting AoIs, the agent is allowed to leave its $m_t$ if necessary to cover a promising area or venture towards an unknown area. When it leaves its current $m_t$, a new $m_{t+1}$ is formed.

AoI exploitation utilizes another neural network, with a different state representation, $\tilde s$, and is trained with the Recurrent-A2C algorithm as showed in ~\ref{fig:network}. $\tilde s$ is a vector of the $o_t$, and the visited information for each adjacent cell in $V_t$. The objective function in Eq.\ref{eq:targets} is triggered after $\kappa_2$ time steps have elapsed, which is a parameter configurable based on the given environment. The AoI exploitation task is considered complete once the function computes that the best region to explore is no longer the agent's current region.


\subsection{Reward Design}
Our reward design is focused on improving adaptive exploration by rewarding the agent for detecting new AoI and visiting new areas of the map. We define the reward for the Target-Directed Exploration, $R_n$, to minimize the time spent to reach the assigned target region $R$ while maximizing the exploration of new AoIs along the way. Given $x$ as the position that the RL algorithm is evaluating:
\begin {equation} \label{eq:rn}
\begin{aligned}
R_{n}= \begin{cases}
    r^{AoI} &  \text{if} \quad p(AoI|x) > \epsilon  \\
    -r^{visited}*V(x) \quad & \text{if} \quad V(x) > 1 \\
    r^{reach_{T_{L}}}\quad & \text {if} \quad x \in T_{L}\\
    r^{reach_R} \quad & \text {if} \quad x \in R\\
\end{cases}
\end{aligned}
\end {equation}

Where $r^{AOI}$ is the weighted reward value for locating AoI, $-r^{visited}$ for revisiting location $x$, $r^{reach_{T_L}}$ for reaching the cell $T_L$, and $r^{reach_R}$ for reaching the target region $R$. We define the reward for the AoI Exploitation agent, $R_e$, so that it maximizes coverage of the total region, with a priority on areas that have higher probability of containing AoI. Thus, $R_e$ is identical to $R_n$ but without the reachability rewards of $r^{reach_{T_{L}}}$ and $ r^{reach_{R}}$ (which are irrelevant in the exploitation context).


\section{Experiments} \label{sec:experiments}
We evaluate our approach against existing techniques for exploration and consider the impacts of the different choices made in our design. We first describe our experimental setup. We describe simulated experiments that measure adaptive exploration via both AoI and total coverage achieved by the UAV. We then present a physical experiment to show the practical feasibility of our model. Finally, we present an ablation study of our approach.

\subsection{Experimental Setup}
We trained our approach on $35$ randomly simulated maps of AoIs of different shapes and distributions. Without retraining the model, we tested it on $20$ new simulated maps, as well as $5$ maps taken from Planet Satellite. This second experiment represents an application in which the UAV must detect illegal gold mining activities in Amazon rainforest in Peru (explained in more detail in the next section). We built a random scenario generator to create the training and testing maps, without overlapping. The maps are structured as $180 \times 180$ cell grids. The scenario generator randomly places the AoI within the boundaries of the map, using a normal distribution with values for the standard deviation and mean based on environment parameters. Testing is done on a set of maps not seen during training. This allows us to study generalization, i.e. how well our learned policies perform in previously unseen environments.
To measure the impact of real noise on our solution, we tested our approach on a real grassy field where the drone searched for plastic tarps, representing AoI.

\textbf{Action Space}. The agent has 5 actions: move forward, move backward, move left, move right, and hover.

\begin{figure}
\centering
\includegraphics [width=0.4\textwidth, height=4 cm]{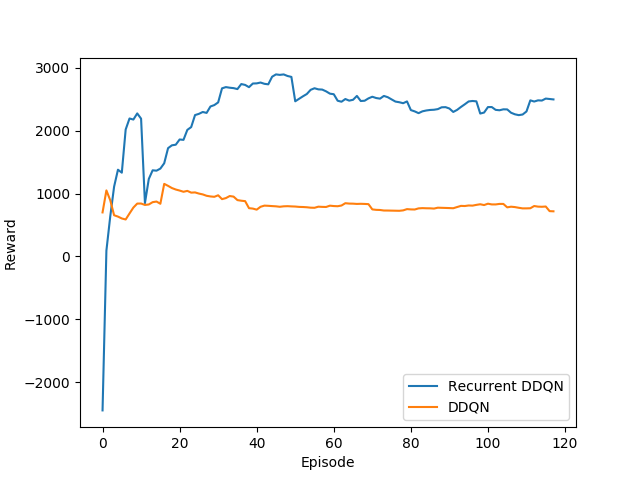}
\includegraphics [width=0.4\textwidth, height=4 cm]{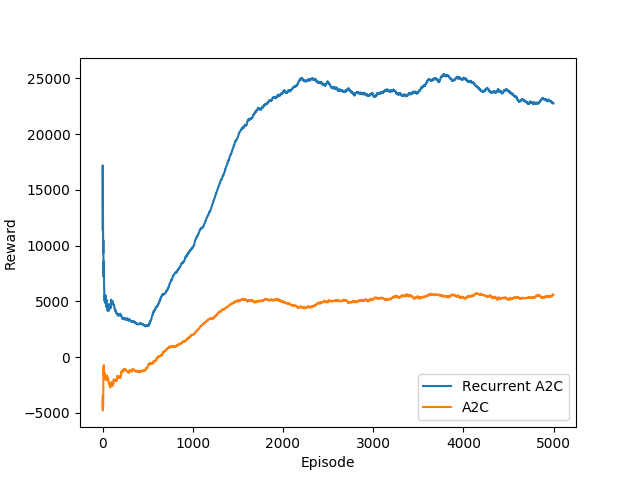}
\caption{Training performance of Recurrent-DDQN vs DDQN in Target-Directed Exploration task (top) and A2C training in AoI Exploitation (bottom)}
\label{fig:training}
\end{figure}

\textbf{Training}. To train the proposed approach, we train each policy separately to excel in its respective task of exploration or exploitation. Then, we leverage transfer learning of DRL to train the entire model, balancing between the two tasks. This effectively speeds up our training process. First, we train the Target-Directed Navigation policy $\pi_n$. This task uses the network architecture in Fig.\ref{fig:network} and Recurrent-DDQN for learning $\pi_{n}$. Since our environment is partially observable, and since a single episode of training can reach hundreds of steps, using Recurrent-DDQN helps the RL agent avoid overestimation and allows it to learn a better policy. We use an input layer with a shape of $[5,31]$, followed by two fully-connected layers with $64$ and $10$ units respectively. The $5$ entries represent the last $5$ time steps, which is important for the LSTM layer to handle the partial observability of the environment. We have an additional input layer for the flattened $m_t$ whose shape is $[5,625]$. That information is processed through two fully-connected layers with $100$ units each, which is then concatenated to the $10$ output units from above. This new vector is the input into another LSTM layer of $110$ units, and then through an output layer of $5$ units, representing the 5 possible actions. We trained this policy with the Adam optimizer, with a learning rate of .001, a batch size of 32, and an experience buffer with a maximum length of 2000. The randomness variable Epsilon in Recurrent-DDQN starts at 0.95 and is multiplied by 0.99 at each update step until it is $<.01$. We also used a discount factor of 0.95. The training results of this task are shown in Fig.\ref{fig:training}. Notice that the inevitable unstable training of DDQN ~\cite{Hausknecht} has been drastically decreased by using the stacked LSTM layers.  

We used Recurrent-A2C  to train Region Exploration because of A2C's ability to solve continuous environments, which policy-gradient methods excel at compared to Q-learning methods. We use the network in Fig. \ref{fig:network}
In the Region Exploration network, the architecture shown is for the actor network, which outputs probabilities for each action. There is also a value network, identical to the actor network, which outputs a single number representing the value assigned to the input. We used the Adam optimizer with a learning rate of .00001 and a batch size of 1. The training results can be seen in Fig.\ref{fig:training}. Once again, we see the significance of included the embedded LSTM stack.

Once the two tasks have been pre-trained separately, to improve how each network interacts with the other, we load the individually-trained network parameters and run 100 full episodes, using the full model architecture shown in Fig.\ref{fig:network}. We then freeze the weights and use them for the testing phase.

\textbf{Baselines}. We tested our approach against 4 baselines;
\begin{enumerate}
    \item \textbf{Sweeping policy} in which the UAV navigates the map in a zigzag fashion. The UAV begins in the top left, travels to the opposite border horizontally, moves down one space, travels back to the left border, and continues on in this fashion without covering any previously covered cell, until the battery threshold is reached;
    \item \textbf{Random policy} which uses the same proposed tasks of Target-Directed exploration and AoI exploitation, but instead of using the DRL policies, randomly picks an action at each time step. This baseline serves to illustrate the success of the DRL networks within our model framework;
    \item \textbf{Curiosity model} which explores the entire map using a curiosity reward function as in \cite{burda2018largescale};
    \item \textbf{Robustness driven exploration} which uses probabilistic metric temporal logic and Monte-Carlo algorithm to build a navigation plan for the UAV in unknown environments. This approach has been introduced in \cite{liu2019robustnessdriven} to navigate Amazonian forests looking for illegal gold mining activities. We compare our results against this approach using the simulated maps for AoI and the illegal mining detection task using the Planet Satellite maps.
\end{enumerate}

\subsection{Testing Results}
We developed a framework for landscape output simulation, which can be used to simulate the output of a neural network on any given landscape image. Like a neural network, this simulation is designed to return a tensor of probabilities for the classification of map features. The map has two classes of features: area of interest (AoI) and non-AoI with adjustable parameters for the density, size, and shape of the AoIs. The simulator can generate maps with the option to add salt-and-pepper noise to the image.
(so that situations like “out of focus” or “lights of reflection” can also be simulated) and to make the AoIs gradients or solids, depending on the desired application. Hence, our simulator can be easily adapted to investigate different types of environments. The developed simulator can be used as an alternative black-box tool for probability tensor outputs that may be used in further tasks \footnote{\url{https://github.com/RL-WFU/Drone_Simulation.git}}.
We simulated $20$ maps of size $180 \times 180$ for the AoI exploration task. We describe the details of the environments used and then analyze the results of our approach against the baselines.
\subsubsection {Task 1: AoI Exploration}
\begin{figure}[h]
\centering
\includegraphics [width=0.42\textwidth, height = 4 cm]{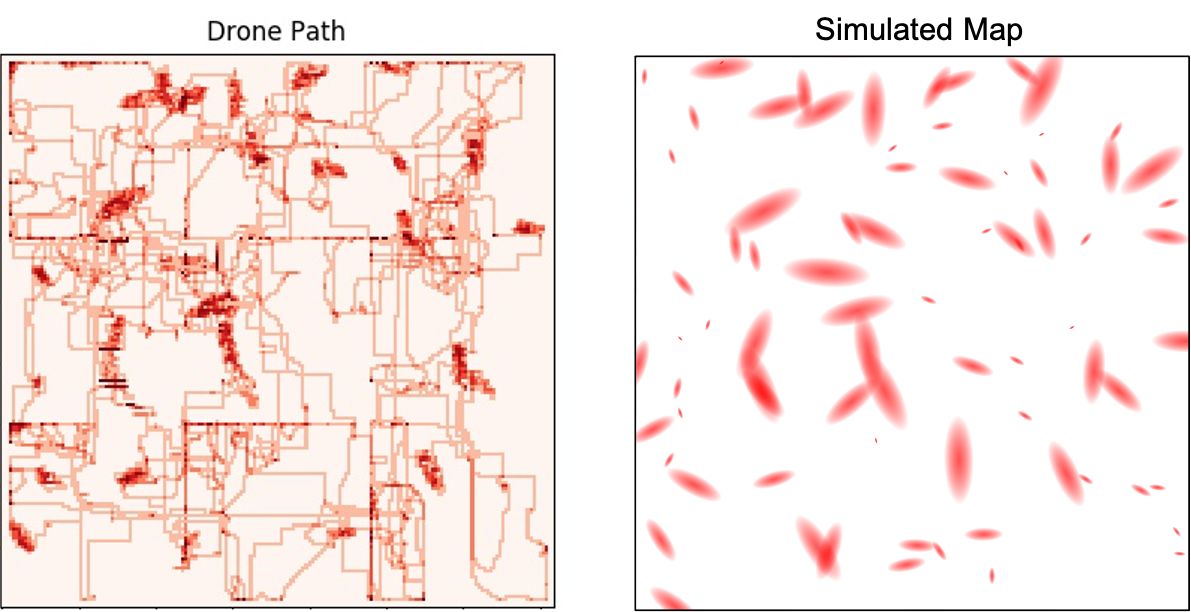}
\caption{Sample path generated by the UAV (left). Darker areas represent places visited more than once. Map simulation (right). Red areas on the map represent abstracted AoIs.}
\label{fig:path}
\end{figure}

For this task, AoI with different shapes and densities are placed randomly across the map. The agent's task is to explore the map efficiently and follow encountered AoI when possible. The agent is given $18,000$ time steps to complete it's mission. We note that the battery threshold of a UAV will be largely dependent on hardware specifications and environment size. Based on the physical time constraints in our drone experiments (discussed further in Task 3), $18000$ is an appropriate upper bound limit. Furthermore, we show results for AoI coverage throughout the episode in order to demonstrate the success of our approach at different time cutoffs (Fig. \ref{fig:episode}).  A representative flight path generated by our model in this task is shown in Fig. \ref{fig:path}. 

Table~\ref{tab:overall} lists the results for our approach and the baselines tested on $20$ maps for this task. Each map was tested $5$ times. Results are averaged across all $100$ tests. Note that the curiosity model achieved the least AoI coverage in every trial because this model must unsystematically explore the map without using smaller, tractable regions. This results in the drone repeatedly exploring the same areas, which significantly reduces overall coverage. Table~\ref{tab:overall} also includes the total map coverage for each model to analyze efficiency. A greater total coverage relative to AoI coverage indicates that more time and resources were spent traversing and imaging irrelevant (i.e. non-AoI) areas. This is particularly notable for the sweeping baseline, which explores near the same amount of the total map as it does specifically AoIs.

In Fig. \ref{fig:episode}, we show the average AoI coverage at each time step during an episode. These results reveal that our model clearly outperforms the baselines in terms of AoI coverage at every point throughout the mission, further indicating its success. This also suggests that a different battery threshold could be considered without significantly altering the results. 

\begin{table*}[t]
    \centering
        \caption{Coverage achieved in 18,000 steps. We report both AoI coverage and total coverage for each model as an indicator of its efficiency. Each value is an average and standard deviation computed from 100 episodes}
    \begin{tabular}{p{0.1\textwidth}||w{l}{0.08\textwidth}w{l}{0.05\textwidth}w{l}{0.07\textwidth}||w{l}{0.07\textwidth}w{l}{0.07\textwidth}w{l}{0.07\textwidth}w{l}{0.07\textwidth}}
    \toprule
    & \multicolumn{3}{c||}{Simulated AoI Detection} &
    \multicolumn{4}{c}{Illegal Mining Detection} 
    \\
    \textit{Model} & &\multicolumn{1}{l}{\textit{Average Coverage}} & \multicolumn{1}{l||}{\textit{}} 
    &
    \textit{ $\;\;$ Average Coverage}
    \\
    \midrule
    \textbf{Our Model} & AoI Coverage: &0.81385 & $\pm$ 0.06740 & \textbf{0.79779} & $\pm$ 0.06462 \\
    & Total Coverage & 0.3051 & $\pm$ 0.03959 & 0.2038 & $\pm$ 0.1394\\
    \hline
   Sweeping & AoI Coverage: & 0.55502 & $\pm$0.07147 & 0.6023 & $\pm$ 0.1936 \\
    & Total Coverage: & 0.5883& $\pm$ 0.1349 & 0.40253 & $\pm$ 0.29538\\
    \hline
    Random & AoI Coverage:& 0.3558 & $\pm$ 0.0653 & 0.1043 & $\pm$ 0.07419 \\
    & Total Coverage: & 0.1708 & $\pm$ 0.04045 & 0.1213 & $\pm$ 0.0893\\
    \hline
    Curiosity~\cite{burda2018largescale} & AoI Coverage: & 0.08192 & $\pm$ 0.0525 & 0.07299 & $\pm$ 0.07833 \\
    & Total Coverage: & 0.03809& $\pm$ 0.01587 & 0.02892 & $\pm$ 0.02504\\
    \hline
    RDE~\cite{liu2019robustnessdriven} & AoI Coverage: & \textbf{0.83807} &$\pm$0.07354 & 0.63522 & $\pm$ 0.05159 \\
    &Total Coverage: & 0.65466&$\pm$0.03254& 0.63530&$\pm$ 0.05145\\
    \bottomrule
    \end{tabular}
    \label{tab:overall}
\end{table*}

\subsubsection {Task 2: Illegal Mining Detection}
We tested our approach on the problem of mapping mercury-based Small-scale Gold Mining (ASGM) in the Amazonian rainforest as described in \cite{liu2019robustnessdriven}. This application is particularly important as ASGM is a significant contributor to deforestation and environmental degradation in the Amazon ~\cite{liu2019robustnessdriven}. Satellite monitoring for ASGM is not possible in cloudy and rainy weather, which is very common in areas like the Amazon rainforest. UAVs can overcome those issues. However, the UAV field of view is significantly smaller than that of a satellite. To use UAVs to collect information in Amazon, acquisition of image data must account for limited flight time, the required storage, and the classification burden of the collected images. In this task, the UAV must explore an area in search of ASGM. The environment is unknown and the only input to the agent is the on-board recognition system. The goal is to maximize the probability of detecting ASGM relative to the exploration effort expended. We tested our approach over five $8\times 8 km^2$ regions in Peru (Delta, Colorado, Madre de Dios, Inambari,La Pampa)~\cite{liu2019robustnessdriven}. We simulated a flight path over these regions using Planet satellite images. The agent's altitude was kept fixed by setting the field of view to $5\times5 m^2$. 

\begin{figure}[h]
\centering
\includegraphics [width=0.4\textwidth, height=8 cm]{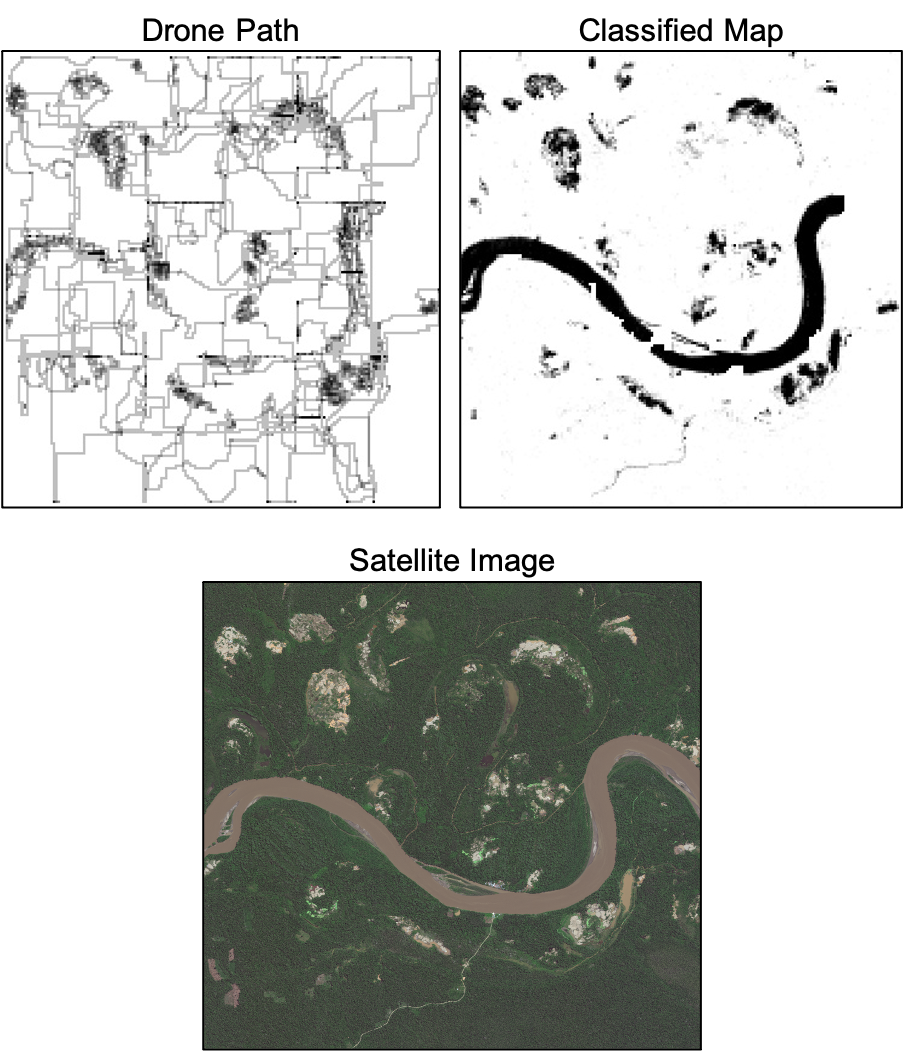}
\caption{Representative sample of the generated path in the illegal mining detection task (top left). Darker areas represent places visited more than once. Map classification (top right). Dark areas represent AoIs related to mining. Satellite map of the area from Amazonian forests in Peru (bottom).}
\label{fig:asgmmap}
\end{figure}

\begin{figure}
\centering
\includegraphics [width=0.45\textwidth]{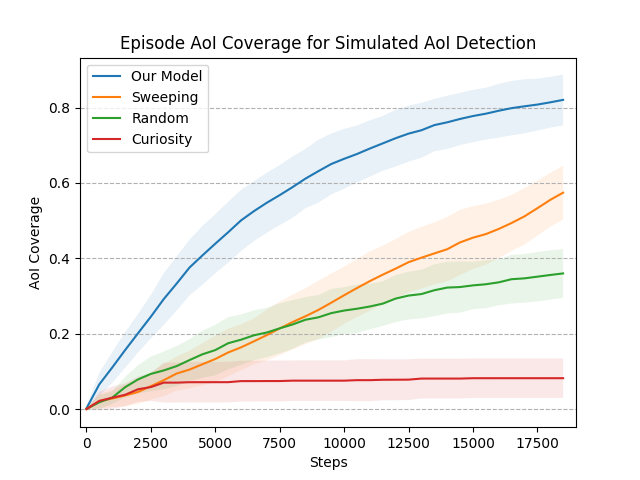}
\includegraphics[width=0.45\textwidth]{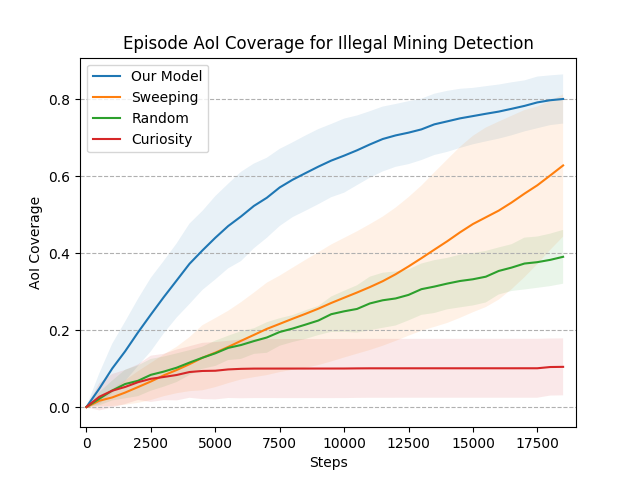}
\caption{AoI coverage achieved by each model throughout a single mission for the simulated AoI exploration task(top) and the mining detection task (bottom).}
\label{fig:episode}
\end{figure}

Table \ref{tab:overall} lists the results for our approach and the baselines tested on $20$ maps for this task. Each map was tested $5$ times, and results were averaged across the $100$ episodes. Our approach explored more mining areas than all other models. It also had a smaller standard deviation, indicating it's consistency and stability relative to the others. We show the AoI coverage collected throughout an episode in Fig. \ref{fig:episode}. On average, our model clearly outperforms the baselines at each time step in this task. Because of the Sweeping policy's high variance, it did achieve similar AoI coverage to our model in some cases. As shown in Table \ref{tab:overall}, the Sweeping policy explored on average 40\% of the whole map in order to cover 60\% of AoI while our model only explored 20\% of the whole map to achieve 79\% AoI coverage. This result demonstrates that, even when our model requires more steps, it more accurately and efficiently navigates the region. Since we do consider the limited storage capacity of the UAV, this is a relevant metric. Furthermore, the high variance of the Sweeping model makes it impractical in real-world deployment. When the UAV happens to begin near the highest concentration of mining AoI, it more quickly covers a larger percent. However, when it happens to begin further away from concentrated mining AoI, it is unable to fully explore it within the given time constraints. Since we assume no prior knowledge about the environment, it is important that a model be able to consistently cover AoI, regardless of its start position. Fig. \ref{fig:episode} indicates that our model is able to do this, while the Sweeping baseline is not. We show the generated paths for one region -- La Pampa -- in Fig. \ref{fig:asgmmap}. The black areas in the map (top right) represent the areas occupied by ASGM while the white represents the forest. Clearly, the agent spent more time exploring ASGM areas and avoided the forest.

\subsubsection {Task 3: Real Experiment}
\begin{figure}
\centering
\includegraphics [width=0.45\textwidth, height= 5 cm ]{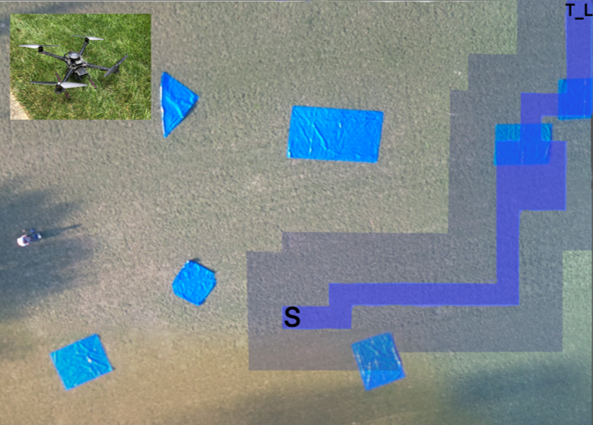}
\caption{The drone path (in purple) from the start position $S$ to its target $T_L$. The blue tarps represent the AoI, and the desaturated purple around the path represents the drone's vision range.
}
\label{fig:realdrone}
\end{figure}

Finally, we deployed our approach in a physical experiment in order to demonstrate its feasibility in a real-world setting. In this test, the drone must explore a soccer field in search of plastic tarps as AoI. The drone used for this task had a Raspberry Pi 3B on-board, equipped with a standard Raspberry Pi camera and image thresholding software to identify colors in HSV format. The testing environment was $50\times50 m^2$ in size. We scale this to correspond to a single egocentric map $m_t$ in our simulations. The drone flew at an altitude of $10 m$. This altitude gives a field of view of approximately $10\times10 m^2$, representing the second level of egocentrism of $o_t$. 
Fig.\ref{fig:realdrone} shows a view of the map $m_t$ for this experiment. The drone's task was to use the Target-Directed Exploration policy $\pi_n$ to travel from its start position $S$ to the top right corner, marked $T_L$. This corresponds to exploring one region and moving to the next in our full simulations. 

As a baseline, we first test the time taken on a straight path between $S$ and $T_L$. With our drone specifications, this task took $1$ minute $8$ seconds on average. Similarly, we determined that the drone can systematically cover the entire map - as in the Sweeping baseline - in $21$ minutes $16$ seconds. We tested our model under the same drone conditions, averaged over $10$ flight missions. The average time to reach $T_L$ using $\pi_n$ was $1$ minute $15$ seconds. The purple path in Fig.\ref{fig:realdrone} represents the drone's path while the shade around the path represents its vision range. Clearly, the drone successfully prioritized the path that had more AoI compared to the straight path to the target. The time taken further indicates that the path generated by our model is still more efficient than traversing every grid. Although this is a simplified experiment compared to our full simulations, it proves the applicability of our approach in real world cases with noisy measurements.

\subsection{Ablation Study}
We performed ablation studies on our approach to pinpoint what parts of our proposed method contribute to the performance. We consider AoI coverage and total map coverage during testing as metrics for comparison. We perform the ablation on the simulated AoI maps from the AoI exploration task.

\subsubsection{Egocentric Map $m$}
\begin{figure}
\centering
\includegraphics [width=0.45\textwidth]{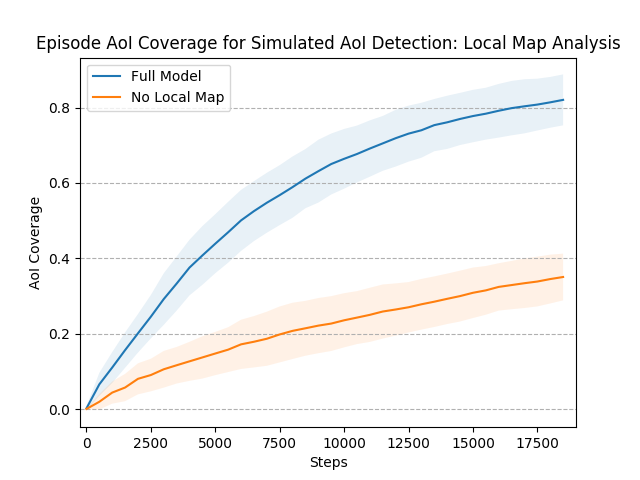}
\caption{Average AoI coverage achieved throughout the episode with each configuration of the egocentric map m.
}
\label{fig:ablation_m}
\end{figure}

We check if the agent's training on the egocentric map $m$ is useful for the adaptive exploration model. We test this by comparing to a model that was not given access to egocentric map $m$ information. Fig.\ref{fig:ablation_m} shows the difference in AoI coverage collected throughout an episode with and without the egocentric map $m$. Training with egocentric maps clearly helps the model achieve significantly higher performance in terms of AoI coverage. The agent trained without access to the egocentric map failed to cover above $40\%$ of the AoI on every map. Table \ref{tab:ablation70} shows the results for AoI coverage after the same $18,000$ time steps from before. The model without the egocentric reached only $34.5\%$ in $18,000$ steps while the egocentric map increased the AoI coverage to $81.4\%$.
\begin{table}[t]
  \caption{Ablation Study: Average map coverage and AoI coverage at the 18,000 step threshold.  Each value is an average and standard deviation computed from 50 episodes}
  \label{tab:ablation70}

  \begin{tabular}{|c|c c|c c|}
    \hline
    \textit{Model} & \textit{AoI Coverage} & & \textit{Total Coverage}  &\\
    \hline
   No V & 0.3214 & $\pm$ 0.0543 & 0.1671 & $\pm$ 0.0169 \\ \hline
   No m	& 0.3449 &	$\pm$ 0.0648 & 0.1401 & $\pm$ 0.0185\\ \hline
   No LSTM & 0.2723	& $\pm$ 0.0861 &0.0802 & $\pm$ 0.0261\\ \hline
   Full Model &	0.8135 & $\pm$ 0.0674 &0.3051 & $\pm$ 0.0396\\ \hline
  \end{tabular}
\end{table}

\subsubsection{Visited Vector $V$}
\begin{figure}
\centering
\includegraphics [width=0.45\textwidth]{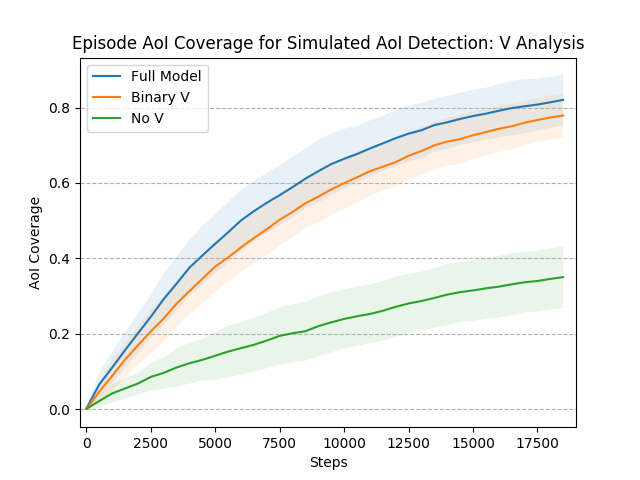}
\caption{ \label{fig:ablation_v} Average AoI coverage achieved throughout the episode with each configuration of the Visited Vector V.
}

\end{figure}

We also study the impact of using the visited vector $V_t$ as an input to the networks for Target-Directed Exploration and AoI Exploitation (Fig.\ref{fig:ablation_v}).  We consider three different levels of comparison: no visited information, binary visited information, and the full $V_t$ we include in our model. As shown in Fig.\ref{fig:ablation_v} and  Table \ref{tab:ablation70}, the model trained without $V_t$ generated exploration policies with notably low AoI coverage throughout the episode and a higher variance from map to map.  Using binary values to represent whether a cell has been visited or not does decrease performance, though we note that this difference is not nearly as significant. Overall, these tests allow us to conclude that including the information about previously visited areas in the training for both $\pi_n$ and $\pi_r$ improves the agent's ability to cover as much of the map as possible while following the AoI when detected. The information from $V_t$ thus helps the agent identify and explore unvisited areas of the map, which pushes the agent towards exploring new areas, ultimately achieving the objective of adaptive exploration.

\subsubsection{Recurrent Layers (LSTM)}
We next investigate the effect of stacking recurrent layers into the nueral networks for both RL algorithms - DDQN and A2C. As before, Table \ref{tab:ablation70} and Fig. \ref{fig:ablation_lstm} reveal that removing the LSTM layers from the model negatively affect its performance in terms of the total coverage and AoI coverage. We refer back to Fig. \ref{fig:training}, which further reveals the importance of the LSTM layers in terms of model training. Overall, the LSTM stack provides the DDQN and A2C networks with a long term memory to "remember" places visited previously in the current episode. Specifically in Target-Directed Exploration, this allow the agent to rationalize its actions in order to decrease distance to the target region with each step. Furthermore, in both tasks the LSTM helps the agent avoid getting stuck in one area, and continue to move to places it hasn't seen before. 

\begin{figure}
\centering
\includegraphics [width=0.45\textwidth]{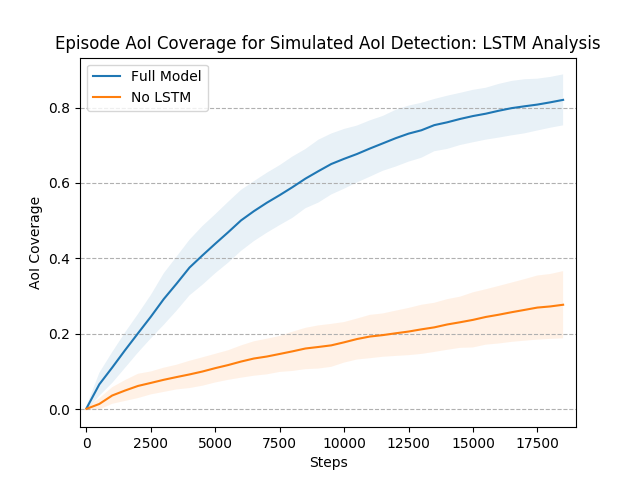}
\caption{ \label{fig:ablation_lstm} Average AoI coverage achieved throughout the episode with and without using an LSTM stack.
}
\label{fig:ablation_lstm}
\end{figure} 


\section{Conclusion} \label{sec:conclusion}

In this paper, we presented an adaptive exploration approach for UAVs searching for areas of interest (AoI) in unknown environments using Deep Reinforcement Learning (DRL). The developed approach started by decomposing the environment into small regions to make this complex problem tractable for RL algorithms. We then employed two deep neural networks, each selecting an action according to one of two navigation tasks: Target-Directed Exploration and AoI Exploitation. The trade-off between these two tasks was controlled using an efficient information gain function which repeatedly computed the best target region to search. We trained the Target-Directed Exploration and AoI Exploitation tasks using Recurrent-DDQN and Recurrent-A2C algorithms, respectively. We tested the proposed approach in 3 different tasks: simulated AoI exploration, ASGM detection, and a real experiment using a drone exploring a field in search of several tarps. We compared our approach results in each task against 4 different baselines from literature. The results showed that our approach outperforms other models in terms of search time, AoI coverage, and overall efficiency. In future work, we intend to add more complexity to the environment by introducing obstacles and adversarial force such as in military settings. We also plan to extend the solution to consider a swarm of UAVs using multi-agent RL algorithms.


\bibliographystyle{ieeetr}
\bibliography{main.bib}

\end{document}